\documentclass{article}
\usepackage{ijcai24}

\usepackage{times}
\usepackage{soul}
\usepackage{url}
\usepackage[hidelinks]{hyperref}
\usepackage[utf8]{inputenc}
\usepackage[small]{caption}
\usepackage{graphicx}
\usepackage{amsmath}
\usepackage{amsthm}
\usepackage{booktabs}
\usepackage{algorithm}
\usepackage{algorithmic}
\usepackage[switch]{lineno}

\usepackage{multirow}
\usepackage{amssymb}
\usepackage{amsmath}
\usepackage{makecell}
\usepackage[table]{xcolor}
\usepackage{color}
\definecolor{lgray}{RGB}{230,230,230}
\title{DCDet: Dynamic Cross-based 3D Object Detector}

\author{
Shuai~Liu
\and
Boyang~Li\and
Zhiyu~Fang\And
Kai~Huang\thanks{Corresponding author.}
\affiliations
School of Computer Science and Engineering, Sun Yat-sen University\\
\emails
\{liush376@mail2, liby83@mail, fangzhy9@mail2, huangk36@mail\}.sysu.edu.cn
}

\begin{document}

\maketitle

\begin{abstract}
Recently, significant progress has been made in the research of 3D object detection. However, most prior studies have focused on the utilization of center-based or anchor-based label assignment schemes. Alternative label assignment strategies remain unexplored in 3D object detection. We find that the center-based label assignment often fails to generate sufficient positive samples for training, while the anchor-based label assignment tends to encounter an imbalanced issue when handling objects with different scales. To solve these issues, we introduce a \textit{dynamic cross label assignment} (DCLA) scheme, which dynamically assigns positive samples for each object from a cross-shaped region, thus providing sufficient and balanced positive samples for training. Furthermore, to address the challenge of accurately regressing objects with varying scales, we put forth a \textit{rotation-weighted Intersection over Union} (RWIoU) metric to replace the widely used $L_1$ metric in regression loss. Extensive experiments demonstrate the generality and effectiveness of our DCLA and RWIoU-based regression loss. The Code is available at \url{https://github.com/Say2L/DCDet.git}.

\end{abstract}

\section{Introduction}

3D object detection plays a crucial role in enabling unmanned vehicles to perceive and understand their surroundings, which is fundamental for ensuring safe driving. Label assignment is a key process for training 3D object detectors. The dominant label assignment strategies in 3D object detection are anchor-based~\cite{pv-rcnn,btcdet,cia-ssd} and center-based \cite{afdet,afdetv2,centerpoint,3dcenternet}. However, both of these label assignment schemes encounter issues that limit the performance of detectors.

\textit{The anchor-based label assignment generally encounters an imbalanced problem when assigning positive samples to objects with different scales.} It employs the prior knowledge of spatial scale for each category to predefine fixed-size anchors on the grid map. By comparing the intersection over union (IoU) between anchors and ground-truth boxes, positive anchors are determined to classify and regress objects. Consequently, the anchor-based label assignment tends to exhibit an uneven distribution of positive anchors across objects of different sizes. For example, car objects typically have a significantly higher number of positive anchors compared to pedestrian objects. This imbalance poses a challenge during training and leads to slow convergence for small objects. Moreover, the anchor-based label assignment scheme necessitates the recalculation of statistical data distribution for different datasets to obtain optimal anchor sizes. This requirement may reduce the robustness of a trained detector when applied to datasets with distinct data distributions.

\textit{The center-based label assignment scheme often faces challenges in providing adequate positive samples for training.} This approach has recently been adopted by various 3D object detectors \cite{afdet,afdetv2,centerpoint,3dcenternet}. It focuses solely on object centers as positive samples (similar to positive anchors). As a result, the number of positive samples remains consistent across objects of different scales, solving the issue of imbalanced positive sample distribution encountered in anchor-based label assignment. However, the center-based label assignment overlooks many potential high-quality positive samples, as only one positive sample per object is responsible for regressing object attributes. This leads to an inefficient utilization of training data and sub-optimal network performance. 

To simultaneously address the aforementioned challenges, this paper introduces a dynamic cross label assignment (DCLA), which aims to provide balanced and ample high-quality positive samples for objects of different scales. Specifically, DCLA dynamically assigns positive samples for each object within a cross-shaped region. The size of this region is determined by a distance parameter, which represents the Manhattan distance from the object's center point. Given the varying scale and potential missing points in point clouds, a dynamic selection strategy is employed to adaptively choose positive samples from the cross-shaped region. As a result, each object is assigned sufficient positive samples, and objects of different scales receive a similar number of positive samples, effectively mitigating the issue of positive sample imbalance.

Moreover, a rotation-weighted IoU (RWIoU) is introduced to accurately regress objects. In the 2D domain, the IoU-based loss \cite{GIoU,DIoU,eiou} is confirmed to be better than the $L_{norm}$ loss. However, in 3D object detection, the development of the IoU-based loss lags behind its 2D counterpart. This challenge arises due to the increased degrees of freedom in the 3D domain. The proposed RWIoU utilizes the idea of rotation weighting, thus elegantly integrating the rotation and direction attributes of objects into the IoU metric. The RWIoU loss can replace the $L_{norm}$ and direction losses to help detectors achieve higher accuracy. Finally, a 3D object detection framework dubbed DCDet is proposed which combines the DCLA and RWIoU. 

The contributions of this work are summarized as follows:
\begin{itemize}
    \item We thoroughly investigate the current widely used label assignment strategies and analyze their pros and cons. Based on experimental observations, we introduce a new label assignment strategy called dynamic cross label assignment (DCLA).
    \item We propose a rotation-weighted IoU (RWIoU) to better measure the proximity of two rotation boxes compared to the $L_1$ metric. RWIoU takes rotations and directions of 3D objects into consideration simultaneously. 
    \item A 3D object detector dubbed DCDet is proposed which combines the DCLA and RWIoU. Extensive experiments on the Waymo Open \cite{waymo} and KITTI \cite{kitti} datasets demonstrate the effectiveness and generality of our methods.
\end{itemize}

\section{Related Work}
\subsection{3D Object Detection}
VoxelNet \cite{voxelnet} encodes voxel features using PointNet \cite{pointnet}, and then extracts features from 3D feature maps through 3D convolutions. SECOND \cite{second} efficiently encodes sparse voxel features by proposed 3D sparse convolution. PointPillars \cite{pointpillars} divides a point cloud into pillar voxels, avoiding the use of 3D convolution and achieving high inference speed. 3DSSD \cite{3dssd} significantly improves inference speed by discarding upsampling layers and refinement networks commonly used in point-based methods. PointRCNN \cite{pointrcnn} produces proposals from raw points using PointNet++ \cite{pointnet++}, and then refines bounding boxes in the second stage. PV-RCNN \cite{pv-rcnn} uses features of internal points to refine proposals. Voxel R-CNN \cite{voxel-rcnn} replaces the features of raw points in the second-stage refinement with 3D voxel features in the 3D backbone.

\subsection{Label Assignment}
Label assignment, which is fundamental to 2D and 3D object detection, significantly influences the optimization of a network. Its development is more mature in 2D object detection, with RetinaNet \cite{focalloss} assigning anchors on the output grid map, FCOS \cite{fcos} designating grid points within the range of ground truth boxes as positive samples, and CenterNet \cite{centernet} identifying center points of ground truth boxes as positive samples. ATSS \cite{atss} and AutoAssign \cite{zhu2020autoassign} propose adaptive strategies for dynamic threshold selection and dynamic positive/negative confidence adjustment, respectively. YOLOX \cite{ge2021yolox} introduces the SimOTA scheme for dynamic positive sample selection. Conversely, 3D object detection label assignment is less developed, grappling with unique challenges such as maintaining a balance of positive samples across various object sizes. Current methods in 3D object detection typically use either anchor-based \cite{second,pointpillars,voxel-rcnn} or center-based \cite{centerpoint,afdet,afdetv2} label assignment schemes. However, these schemes have drawbacks: the anchor-based label assignment often results in unbalanced assignments, and the center-based label assignment may overlook high-quality samples. To simultaneously overcome the above two drawbacks, we propose the dynamic cross label assignment (DCLA). Details about the DCLA are described in the methodology section.

\begin{figure}[t]
    \centering
    \includegraphics[width=0.85\columnwidth]{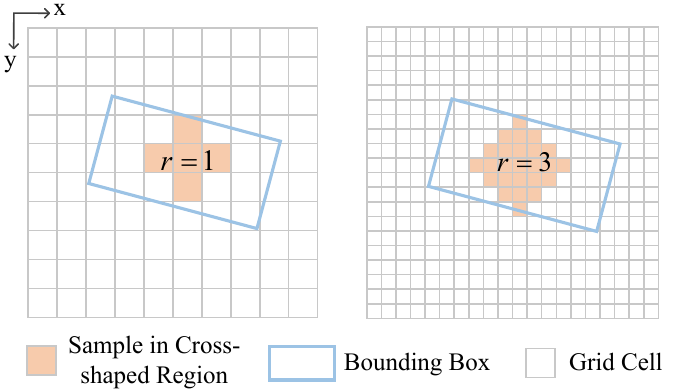}
    \caption{Cross-shaped region for different grid cell sizes. }
    \label{fig1}
\end{figure}

\begin{figure*}[t]
    \centering
    \includegraphics[width=0.9\textwidth]{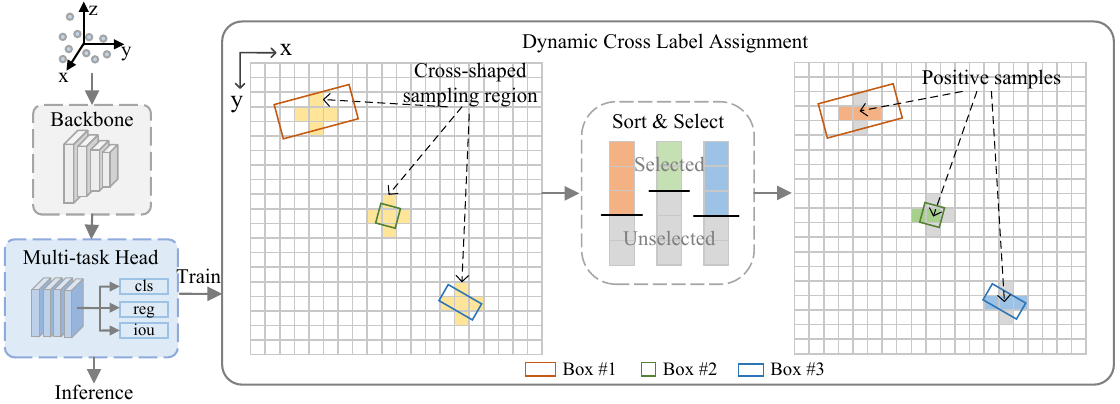}
    \caption{The overall framework of our DCDet. The dynamic cross label assignment scheme is only used in the training phase. }
    \label{fig2}
\end{figure*}

\subsection{IoU-based Loss}
IoU-based losses \cite{GIoU,DIoU,eiou} without rotation have been well studied in 2D object detection. These methods not only ensure consistency between the training objective and the evaluation metric but also normalize object attributes, leading to enhanced performance compared to the $L_{norm}$ loss. Due to their success in 2D object detection, some 3D object detection methods \cite{3diou,rdiou,pillarnet} incorporate IoU-based losses. 3DIoU \cite{3diou} extends IoU calculation from 2D to 3D by considering rotation. However, the optimization direction of 3DIoU-based loss can be opposite to the correct direction. To address this, RDIoU \cite{rdiou} decouples rotation from 3DIoU. It considers rotation as an attribute similar to object location, but it doesn't consider object direction. A direction loss is needed for classifying object directions. ODIoU \cite{pillarnet} combines $L_1$ metric and axis-aligned IoU to regress objects. Our proposed RWIoU incorporates both rotation and direction into the IoU metric, eliminating the need for $L_{norm}$ and direction losses. Details of RWIoU will be explained in the next section.

\section{Methodology}
This section will describe the dynamic cross label assignment (DCLA) and the rotation-weighted IoU (RWIoU) in detail. The overall framework is illustrated in Figure~\ref{fig2}.

\subsection{Dynamic Cross Label Assignment}
The label assignment schemes used in existing 3D object detection methods are generally based on prior information such as spatial ranges or object scales to manually select positive samples. For example, the anchor-based label assignment uses object scales to set the sizes of anchors and then uses anchors with IoU greater than a certain threshold as positive samples. The anchor-based label assignment generally produces unbalanced positive samples for different-scale objects, causing the model to prioritize large-scale objects. The center-based label assignment usually takes the center points of grounding truths as positive samples. This can result in a large number of good-quality samples being discarded, leaving inefficient utilization of training data. 

The above label assignment schemes have a common property, they all use static prior information as the selection criteria. And the prior information is determined by human experience. Dynamic label assignment schemes \cite{atss,zhu2020autoassign,ge2021yolox} have shown their advantages in 2D object detection. However, directly transferring these schemes to 3D object detection is not trivial. There are some challenges: 1) There is no space to dynamically select positive samples for small objects (e.g. pedestrians). Because small objects generally cover one or two grid points on the output map; 2) The coverage of objects with different scales varies greatly. This easily results in an imbalance of positive samples between different scale objects. 

To dynamically select sufficient high-quality positive samples while maintaining the balance between different scale objects, we propose a dynamic cross label assignment (DCLA) scheme. Specifically, it limits the positive sampling range in a cross-shape region for each object. Typically, an object's center region on a feature map contains enough features to identify it \cite{fcos}, and objects in point clouds have regular shapes. Therefore, we only use the center point and its surrounding points for positive sampling in the DCLA scheme. We refer to this sampling range as the cross region. It can be adjusted by a parameter $r$ to adapt to outputs with different grid cell sizes as illustrated in Figure~\ref{fig1}. $r$ is the Manhattan distance away from the center point. When $r=1$, the cross region covers the center and its top, down, left, and right neighbors. And when $r=0$, the DCLA degenerates to the center-based label assignment.

The implementation steps of the DCLA are described in detail next. Given a ground truth $\mathbf{b}^{t}$ and positions $P$ in its cross region, calculate the selection cost as follows:

\begin{equation}
    c_{j} = L_{j}^{cls} + \lambda_{reg} L_{j}^{reg}, j \in P,
\label{Eq:eq1}
\end{equation}

\noindent where the $L_{j}^{cls}$ and $L_{j}^{reg}$ are the classification loss and regression loss between the ground truth $\mathbf{b}^{t}$ and $j$-th prediction $\mathbf{b}^{o}_{j}$ respectively, and $\lambda_{reg}$ is the weight of regression loss. Then, sort the predictions in the cross region according to the selection costs. Next, sum the IoUs between the ground truth $\mathbf{b}^{t}_{i}$ and predictions $\mathbf{b}^{o}_{j}, j \in P$:

\begin{equation}
    k = \max(\lfloor \sum_{j \in P} \mathrm{IoU}(\mathbf{b}^{t}, \mathbf{b}^{o}_{j})\rfloor, 1).
\label{Eq:eq2}
\end{equation}

\begin{figure*}[t]
    \centering
    \includegraphics[width=0.8\textwidth]{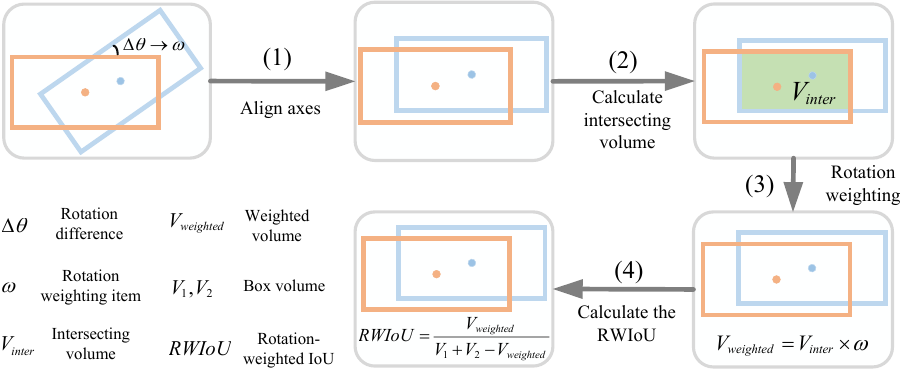}
    \caption{The calculation process of RWIoU.}
    \label{fig3}
\end{figure*}

\noindent We utilize $k$ as the number of positive samples for ground truth $\mathbf{b}^{t}$. Finally, select the top $k$ predictions as positive samples. And the rest predictions are negative samples.

%Specifically, given a point cloud input and ground truth boxes $\{\mathbf{b}_1^t, \mathbf{b}_2^t, \cdots , \mathbf{b}_n^t\}$. Assuming $f(\mathbf{b}^{t}_i, \mathbf{b}^{o}_{ij})$ is the regression loss function for each predicted box, where $\mathbf{b}^{t}_i$ and $\mathbf{b}^{t}_{ij}$ are the $i$-th ground truth box and its $j$-th predicted box, respectively. Thus the regression loss for a point cloud is calculated as follows:

Specifically, given a point cloud input and the ground truth boxes $\{\mathbf{b}_1^t, \mathbf{b}_2^t, \cdots, \mathbf{b}_n^t\}$, we assume that $f(\mathbf{b}^{t}_i, \mathbf{b}^{o}_{ij})$ represents the regression loss function, where $\mathbf{b}^{t}_i$ and $\mathbf{b}^{o}_{ij}$ denote the $i$-th ground truth box and its $j$-th predicted box, respectively. Therefore, the regression loss $\ell$ for the point cloud is calculated as follows:

\begin{equation}
    \begin{aligned}
        \ell &= \frac{1}{N} \sum_{i=1}^{n} \sum_{j=1}^{k_i}f(\mathbf{b}^{t}_i, \mathbf{b}^{o}_{ij}), \\
        N &= \sum_{i=1}^{n} \sum_{j=1}^{k_i}1,
    \end{aligned}
\label{Eq:eq3}
\end{equation}

\noindent where $N$ represents the total number of positive samples in the input point cloud, and $k_i$ denotes the number of positive samples assigned for the ground truth $\mathbf{b}^{t}_{i}$. Notably, $k_i$ is calculated independently for each ground truth, as in Eq.~(\ref{Eq:eq2}). It is related to the number of high-quality samples in the cross region and is not dependent on the ground truth scale. However, in the anchor-based label assignment, $k_i$ varies significantly with the ground truth scale, resulting in a bias towards large-scale objects in the loss. For the center-based label assignment, $k_i$ is always equal to 1, leading to inefficient utilization of training data.

We adopt the heatmap target for the classification task. The weights of positive samples are set to 1, and the weights of negative samples in cross regions are set to the values of IoU between predicted boxes and ground-truth boxes. As for the rest negative samples, the weights are all set to 0.

%It is worth noting that the $k$ is computed independently for different ground truths, so the more high-quality samples within the range of a certain true frame, the larger the value, and the more positive samples for that true frame. This dynamic calculation of values allows the model to focus on optimizing high-quality samples, thus accelerating model convergence.

\subsection{Rotation-Weighted IoU}
%In general, the scales of different category objects vary greatly, and different attributes (e.g., location, size, and rotation) of an object also have scale differences. Most existing methods use the $L_{norm}$ loss as the regression loss. However, the $L_{norm}$ loss makes the model vulnerable to the differences in object scales as well as attribute scales. Thus, the loss of large objects and attributes will account for most of the total loss. The IoU metric can normalize the attributes of objects, so it is not affected by the scale differences. In addition, the optimization objective of the IoU-based loss is consistent with the evaluation metrics of detection models, so using the IoU-based loss instead of the $L_{norm}$ loss often leads to accuracy improvement \cite{DIoU,GIoU,rdiou}.

In general, different object categories exhibit significant scale variations, and various attributes such as location, size, and rotation also possess scale differences. Many existing methods employ the $L_{norm}$ loss as the regression loss. However, this loss function renders the model sensitive to differences in both object and attribute scales. Consequently, large objects and attributes dominate the total loss. The IoU metric can normalize object attributes, making it immune to scale differences. Moreover, the optimization objective of the IoU-based loss aligns with the evaluation metrics of detection models. Therefore, substituting the $L_{norm}$ loss with the IoU-based loss often yields accuracy improvement \cite{DIoU,GIoU,rdiou}.

%Using IoU-based loss in 3D object detection presents several obstacles. First, calculating IoU requires the computation of the polyhedron volume, which is complex and computationally expensive. Second, due to the rotation being tightly coupled in IoU calculation, the traditional IoU-based loss can misdirect the optimization, causing instability in training \cite{rdiou}. Lastly, the traditional IoU metric is difficult to integrate with the directions of objects. Therefore $L_1$ loss or direction loss is needed to assist models in classifying the directions. 

Utilizing IoU-based loss in 3D object detection poses several challenges. Firstly, calculating traditional IoU requires the computation of polyhedron volumes, which is a complex and computationally expensive task. Secondly, the traditional IoU-based loss, due to its tight coupling with rotation, can lead to misdirection in optimization, resulting in training instability \cite{rdiou}. Lastly, integrating the traditional IoU metric with object directions is not trivial. Therefore, the inclusion of $L_1$ loss or direction loss becomes necessary to aid models in classifying object directions.

To tackle the aforementioned challenges, we propose a rotation-weighted IoU (RWIoU). It thoroughly decouples the rotation from the IoU calculation, making the computation similar to the axis-aligned IoU computation. RWIoU can be implemented with just a few lines of code. By integrating sine and cosine values of rotations of objects into a rotation weighting item, our RWIoU can penalize rotation and direction errors simultaneously. 

The RWIoU calculation process is shown in Figure~\ref{fig3}. It first considers two rotation boxes $\mathbf{B_1}$ and $\mathbf{B_2}$ as axis-aligned boxes, and then calculates the intersecting volume of the two axis-aligned boxes as follows:

\begin{equation}
    \begin{aligned}
    s_{L}  = & \max \left(x_{1}-l_{1} / 2, x_{2}-l_{2} / 2\right), \\
    s_{R} = &\min \left(x_{1}+l_{1} / 2, x_{2}+l_{2} / 2\right), \\
    s_{B} = &\max \left(y_{1}-w_{1} / 2, y_{2}-w_{2} / 2\right), \\
    s_{T} = &\min \left(y_{1}+w_{1} / 2, y_{2}+w_{2} / 2\right), \\
    s_{D} = &\max \left(z_{1}-h_{1} / 2, z_{2}-h_{2} / 2\right), \\
    s_{U} = &\min \left(z_{1}+h_{1} / 2, z_{2}+h_{2} / 2\right), \\
    V_{\text {inter }} = & \max \left(s_{R}-s_{L}, 0\right) \times \max \left(s_{T}-s_{B}, 0\right) \\
    & \times \max \left(s_{U}-s_{D}, 0\right),
    \end{aligned}
    \label{Eq:eq4}
\end{equation}

\noindent where $(x_i, y_i, z_i), i\in \{1,2\}$ denote the locations of box centers, $(l_i, w_i, h_i), i\in \{1,2\}$ represent the sizes of boxes, and $V_{inter}$ denotes the intersecting volume of two axis-aligned boxes. Then, we update the $V_{inter}$ according to the rotation difference of the two boxes as follows:

\begin{equation}
    \begin{aligned}
        V_{weighted} = & \omega V_{inter}, \\
        \omega = & \omega_{s} \omega_{c},\\
        \omega_{s} = &(1 - \alpha \frac{|\textrm{sin}\theta_{2} - \textrm{sin}\theta_{1}|}{2}),\\
        \omega_{c} = &(1 - \alpha \frac{|\textrm{cos}\theta_{2} - \textrm{cos}\theta_{1}|}{2}),\\
    \end{aligned}
    \label{Eq:eq5}
\end{equation}

\noindent where $\theta_{1}$ and $\theta_{2}$ represent rotations of two boxes, $\omega_s$ and $\omega_c$ denote the sine and cosine rotation error factor respectively that are all normalized to the range of $[0, 1]$, $\omega$ represents the rotation weighting item,  $V_{weighted}$ is the rotation-weighted value of $V_{inter}$, $\alpha \in \left[0, 1\right]$ is a hyper-parameter which is used to control the contribution of rotation to the RWIoU. If $\alpha=0$, the RWIoU degrades to axis-aligned IoU. After obtaining $V_{weighted}$, the value of RWIoU can be calculated as follows:

\begin{equation}
    \begin{aligned}
    V_{\text {union }} & =V_{1}+V_{2}-V_{\text {weighted }}, \\
    RWIoU & =\frac{V_{\text {weighted }}}{V_{\text {union }}},
\end{aligned}
    \label{Eq:eq6}
\end{equation}

\noindent where $V_{1}$ and $V_{2}$ represent the volumes of two boxes, respectively. The gradient analysis of the RWIoU is in Appendix.  

\begin{table*}[t]
    \centering
    \resizebox{\linewidth}{!}{
    \begin{tabular}{cccccccccc}
    \hline
    \multirow{2}{*}{Method} & \multirow{2}{*}{Stages} & {LEVEL 2} & \multicolumn{3}{c}{LEVEL 1} && \multicolumn{3}{c}{LEVEL 2} \\
    \cline{4-6} \cline{8-10}
    & & mAP/mAPH & Vehicle & Pedestrian & Cyclist && Vehicle & Pedestrian & Cyclist\\
    \hline
    LiDAR R-CNN (a) \cite{lidarrcnn} & 2 & 65.8/61.3 & 76.0/75.5 & 71.2/58.7 & 68.6/66.9 && 68.3/67.9 & 63.1/51.7 & 66.1/64.4\\
    Part-A2-Net (a) \cite{parta2} & 2 & 66.9/63.8 & 77.1/76.5 & 75.2/66.9 & 68.6/67.4 && 68.5/68.0 & 66.2/58.6 & 66.1/64.9 \\
    Voxel R-CNN$\dagger$ (a) \cite{voxel-rcnn} & 2 & 68.6/66.2 & 76.1/75.7 & 78.2/72.0 & 70.8/69.7 && 68.2/67.7 & 69.3/63.6 & 68.3/67.2 \\
    PV-RCNN$\dagger$ (c) \cite{pv-rcnn} &  2 & 69.6/67.2 & 78.0/77.5 & 79.2/73.0 & 71.5/70.3 &&  69.4/69.0 & 70.4/64.7 & 69.0/67.8 \\
    PV-RCNN++$\dagger$ (c) \cite{pv-rcnn++} & 2 & 71.7/69.5 & 79.3/78.8  & 81.8/76.3 & 73.7/72.7 && 70.6/70.2 & 73.2/68.0 & 71.2/70.2 \\
    FSD \cite{fsd} & 2 & 72.9/70.8 & 79.2/78.8 & 82.6/77.3 & 77.1/76.0 && 70.5/70.1 & 73.9/69.1 & 74.4/73.3 \\
    \hline
    SECOND* (a) \cite{second} & 1 & 61.0/57.2 & 72.3/71.7 & 68.7/58.2 & 60.6/59.3 && 63.9/63.3 & 60.7/51.3 & 58.3/57.0 \\
    PointPillars* (a) \cite{pointpillars} & 1 & 62.8/57.8 & 72.1/71.5 & 70.6/56.7 & 64.4/62.3 && 63.6/63.1 & 62.8/50.3 & 61.9/59.9 \\
    IA-SSD (a) \cite{iassd} & 1 & 66.8/63.3 & 70.5/69.7 & 69.4/58.5 & 67.7/65.3 && 61.6/61.0 & 60.3/50.7 & 65.0/62.7 \\
    SST* (a) \cite{sst} & 1 & 67.8/64.6 & 74.2/73.8 & 78.7/69.6 & 70.7/69.6 && 65.5/65.1 & 70.0/61.7 & 68.0/66.9\\
    CenterPoint$\ddagger$ (c) \cite{centerpoint} & 1 & 68.2/65.8 & 74.2/73.6 & 76.6/70.5 & 72.3/71.1 && 66.2/65.7 & 68.8/63.2 & 69.7/68.5 \\
    VoxSet (c) \cite{voxset} & 1 & 69.1/66.2 & 74.5/74.0 & 80.0/72.4 & 71.6/70.3 && 66.0/65.6 & 72.5/65.4 & 69.0/67.7 \\
    PillarNet (c) \cite{pillarnet} & 1 & 71.0/68.5 & 79.1/78.6 & 80.6/74.0 & 72.3/66.2 && 70.9/70.5 & 72.3/66.2 & 69.7/68.7\\
    AFDetV2 (c) \cite{afdetv2} & 1 & 71.0/68.8 & 77.6/77.1 & 80.2/74.6 & 73.7/72.7 && 69.7/69.2 & 72.2/67.0 & 71.0/70.1 \\
    CenterFormer (c) \cite{centerformer} & 1 & 71.1/68.9 & 75.0/74.4 & 78.6/73.0 & 72.3/71.3 &&  69.9/69.4 & 73.6/68.3 & 69.8/68.8 \\
    SwinFormer (c) \cite{swformer} & 1 & -/- & 77.8/77.3 & 80.9/72.7 & -/- && 69.2/68.8 & 72.5/64.9 & -/- \\
    PillarNeXt (c) \cite{pillarnext} & 1 & 71.9/69.7 & 78.4/77.9 & 82.5/77.1 & 73.2/72.2 && 70.3/69.8 & 74.9/69.8 & 70.6/69.6\\
    DSVT (Pillar) (c) \cite{dsvt} & 1 & 73.2/71.0 & 79.3/78.8 & 82.8/77.0 & 76.4/75.4 && 70.9/70.5 & 75.2/69.8 & 73.6/72.7 \\
    %DSVT (Voxel) (c) \cite{dsvt} & 1 & 74.0/72.1 & \textbf{79.7/79.3} & 83.7/78.9 & 77.5/76.5 & 71.4/71.0 & 76.1/71.5 & 74.6/73.7 \\
    \hline
    DCDet (20\%) (ours) & 1 & 74.0/71.5 & 79.2/78.7 & 83.8/77.6 & 77.4/76.3 && 71.0/70.6 & 76.2/70.2 & 74.8/73.7\\
    DCDet (ours) & 1 & \textbf{75.0/72.7} & \textbf{79.5/79.0} & \textbf{84.1/78.5} & \textbf{79.4/78.3} && \textbf{71.6/71.1} & \textbf{76.7/71.3} & \textbf{76.8/75.7}\\
    \hline
    \end{tabular}
    }
    \caption{Performance comparisons on the Waymo Open validation set. The results of AP/APH are reported. *: reported by \protect\cite{fsd}. $\dagger$: reported by \protect\cite{pv-rcnn++}. $\ddagger$: reported by \protect\cite{dsvt}. `a' and `c' denote the anchor-based and center-based label assignment, respectively. `20\%' denotes only 20\% training samples are used. }
\label{table1}
\end{table*}

\begin{table*}[t]
    \centering
    \resizebox{\linewidth}{!}{
    \begin{tabular}{ccccccccc}
    \hline
    \multirow{2}{*}{Method} & {LEVEL 2} & \multicolumn{3}{c}{LEVEL 1} && \multicolumn{3}{c}{LEVEL 2} \\
    \cline{3-5} \cline{7-9}
    & mAP/mAPH & Vehicle & Pedestrian & Cyclist && Vehicle & Pedestrian & Cyclist\\
    \hline
    CenterPoint \cite{centerpoint} & - & 80.2/79.7 & 78.3/72.1 & - && 72.2/71.8 & 72.2/66.4 & - \\
    PV-RCNN \cite{pv-rcnn} & 71.2/68.8 & 80.6/80.2 & 78.2/72.0 & 71.8/70.4  && 72.8/72.4 & 71.8/66.1 & 69.1/67.8 \\
    PillarNet-18 \cite{pillarnet} & 71.3/68.5 & 81.9/81.4 & 80.0/72.7 & 68.0/66.8 && 74.5/74.0 & 74.0/67.1 & 65.5/64.4\\
    AFDetV2 \cite{afdetv2} & 72.2/70.0 & 80.5/80.0 & 79.8/74.4 & 72.4/71.2 && 73.0/72.6 & 73.7/68.6 & 69.8/68.7\\
    PV-RCNN++ \cite{pv-rcnn++} & 72.4/70.2 & 81.6/81.2 & 80.4/75.0 & 71.9/70.8 && 73.9/73.5 & 74.1/69.0 & 69.3/68.2 \\
    DCDet (ours) & \textbf{75.7/73.3} & \textbf{82.2/81.7} & \textbf{83.4/77.8} & \textbf{77.3/76.1} && \textbf{74.8/74.4} & \textbf{77.5/72.1} & \textbf{74.7/73.5}\\
    \hline
    \end{tabular}
    }
    \caption{Performance comparisons on the Waymo Open test set by submitting to the official test evaluation server. The results are achieved by using single point cloud frames. No test-time augmentations are used. }
\label{table6}
\end{table*}

\subsection{Loss Function}
Single-stage detectors typically encounter misalignment between classification confidence and localization accuracy. To solve the misalignment problem, we follow \citeauthor{cia-ssd} to introduce an extra IoU prediction branch. The classification loss $L_{cls}$ and IoU prediction loss $L_{iou}$ are the same as those of CIA-SSD \cite{cia-ssd}.

The regression loss $L_{reg}$ is based on the RWIoU. It is calculated as follows:

\begin{equation}
    \begin{aligned}
    L_{reg} = & \frac{1}{N} \sum_{i=1}^{N} 1-R W I o U_{i}+(\frac{D_{i}}{Diag_{i}})^{2},
    \end{aligned}
    \label{Eq:eq7}
\end{equation}

%\begin{figure}[t]
%    \centering
%    \includegraphics[width=0.8\columnwidth]{2024-01-09_16-47.png}
%    \caption{Comparison of learning behaviors between different models. The results of LEVEL 2 APH on the Waymo Open validation set are reported (10\% training data). }
%    \label{fig4}
%\end{figure}

%\noindent where $n$ is total number of positive samples, $RWIoU_i$ and $D_{i}$ are the RWIoU value and  the L2 distance of centers between the $i$-th predicted and ground truth boxes, and $Diag_{i}$ is the diagonal of the minimal enclose rectangle of the $i$-th predicted and ground truth boxes. The item $\frac{D_{i}}{Diag_{i}}$ is used to optimize the prediction of center locations. Since our RWIoU employs sine and cosine functions to represent the rotation angle of a bounding box, the direction loss is no longer required. Finally, the overall loss function is calculated as follows:

\noindent where $N$ is the total number of positive samples, $RWIoU_i$ and $D_{i}$ represent the RWIoU value and the $L_2$ distance of centers, respectively. Additionally, $Diag_{i}$ denotes the diagonal length of the minimal enclosing rectangle of the $i$-th predicted box and its ground truth. The term $\frac{D_{i}}{Diag_{i}}$ is used to optimize the prediction of center locations. Since our RWIoU incorporates sine and cosine functions to represent the rotation angle of a bounding box, the need for a direction loss is eliminated. The overall loss function is calculated as follows:

\begin{equation}
    L = \lambda_{cls}L_{cls} + \lambda_{reg} L_{reg} + \lambda_{iou}L_{iou},
\label{Eq:eq8}
\end{equation}

\noindent where $\lambda_{cls}$, $\lambda_{reg}$, and $\lambda_{dir}$ are the weight of classification, regression, and direction losses, respectively.

\section{Experiments}
In this section, we evaluate models on widely-used 3D object detection benchmark datasets including Waymo Open \cite{waymo} and KITTI \cite{kitti}. 
%Furthermore, extensive ablation experiments are conducted to validate the effectiveness of the proposed DCLA scheme and RWIoU.

\begin{table*}[t]
    \centering
    \resizebox{\linewidth}{!}{
    \begin{tabular}{ccccccccccc}
    \hline
    \multirow{2}{*}{Method} & \multirow{2}{*}{Training Data} &\multicolumn{4}{c}{LEVEL 1} && \multicolumn{4}{c}{LEVEL 2} \\
    \cline{3-6} \cline{8-11}
    && mAP/mAPH & Vehicle & Pedestrian & Cyclist && mAP/mAPH & Vehicle & Pedestrian & Cyclist\\
    \hline
    SECOND & 20\% & 64.8/60.4 &70.9/70.3 & 65.8/54.8 & 57.8/56.2 && 58.7/54.7 & 62.6/62.0 & 57.8/48.0 & 55.7/54.2\\
    SECOND* & 20\% & 73.4/70.0 &74.0/73.3 & 77.0/69.1 & 69.2/67.7 && 67.1/64.0 & 65.7/65.2 & 68.7/61.3 & 66.9/65.4\\
    \rowcolor{lgray} Improvement $\uparrow$ & N/A & +8.6/+9.6 & +3.1/+3.0 & +11.2/+14.3 & +11.4/+11.5 && +8.4/+9.3 & +3.1/+3.2 & +10.9/+13.3 & +11.2/+11.2\\
    \hline
    PillarNet & 20\% & 71.6/68.0 & 72.9/72.3 & 73.0/64.1 & 68.9/67.6 && 65.6/62.3 & 64.9/64.4 & 65.3/57.2 & 66.5/65.2\\
    PillarNet* & 20\% &75.1/70.9 & 75.6/75.0 & 78.1/67.7 & 71.7/70.0 && 69.0/65.1/ & 67.8/67.3 & 70.0/60.4 & 69.2/67.6\\
    \rowcolor{lgray} Improvement $\uparrow$ & N/A & +3.5/+2.9 & +2.7/+2.7 & +5.1/+3.6 & +2.8/+2.4 && +3.4/+2.8 & +2.9/+2.9 & +4.7/+3.2 & +2.7/+2.4\\
    \hline
    DSVT & 20\% & 78.3/75.3 & 78.1/77.6 & 82.3/74.8 & 74.6/73.5 && 72.2/69.3 & 69.8/69.3 & 74.7/67.7 & 72.0/71.0\\
    DSVT* & 20\% & 79.8/76.5 & 79.2/78.7 & 83.6/75.3 & 76.5/75.4 && 73.7/70.6 & 71.1/70.7 & 76.2/68.3 & 73.9/72.8\\
    \rowcolor{lgray} Improvement $\uparrow$ & N/A &+1.5/+1.2 & +1.1/+1.1 & +1.3/+0.5 & +1.9/+1.9 && +1.5/+1.3 & +1.3/+1.4 & +1.5/+0.6 & +1.9/+1.8\\
    \hline
    SECOND & 100\% & 67.2/63.1 & 72.3/71.7 & 68.7/58.2 & 60.6/59.3 && 61.0/57.2 & 63.9/63.3 & 60.7/51.3 & 58.3/57.1 \\
    SECOND*& 100\% & 74.2/71.0 & 74.4/73.8 & 78.4/70.8 & 69.9/68.5 && 68.0/65.1 & 66.3/65.9 & 70.2/63.2 & 67.5/66.1\\
    \rowcolor{lgray} Improvement $\uparrow$ & N/A & +7.0/+7.9 & +2.1/+2.1 & +9.7/+12.6 & +9.3/+9.2 && +7.0/+11.9 & +2.4/+2.6 & +9.5/+12.9 & +9.2/+9.0\\
    \hline
    PillarNet & 100\% & 73.4/70.0 & 74.0/73.5 & 75.3/66.9 & 70.8/69.6 && 67.4/64.3 & 66.2/65.7 & 67.7/60.0 & 68.3/67.1\\
    PillarNet* & 100\% & 75.7/71.9 & 75.8/75.3 & 79.1/69.7 & 72.2/70.7 && 69.7/66.1 & 68.2/67.6 & 71.1/62.4 & 69.8/68.4\\
    %PillarNet* & 73.0/69.7 & 73.3/72.7 & 75.4/67.3 & 70.3/69.1 & 66.5/63.5 & 65.0/64.4 & 66.8/59.3 & 68.0/66.8\\
    \rowcolor{lgray} Improvement $\uparrow$ & N/A & +2.3/+1.9 & +1.8/+1.8 & +3.8/+2.8 & +1.4/+1.1 && +2.3/+1.8 & +2.0/+1.9 & +3.4/+2.4 & +1.5/+1.3\\
    \hline
    %DSVT & 67.2/64.2 & 73.5/73.0 & 62.3/54.8 & 65.9/64.8 & 60.6/57.9 & 64.9/64.4 & 53.5/47.0 & 63.5/62.4\\
    DSVT & 100\% & 80.1/77.4 & 79.1/78.6 & 82.7/76.3 & 78.4/77.3 && 73.8/71.3 & 70.9/70.5 & 75.0/68.9 & 75.6/74.6\\
    DSVT* & 100\% & 81.5/78.7 & 80.4/79.9 & 84.5/77.4 & 79.7/78.6 && 75.7/72.9 & 72.6/72.1 & 77.2/70.4 & 77.2/76.2\\
    \rowcolor{lgray} Improvement $\uparrow$ & N/A & +1.4/+1.3 & +1.3/+1.3 & +1.8/+1.1 & +1.3/+1.3 && +1.9/+1.6 & +1.7/+1.6 & +2.2/+1.5 & +1.6/+1.6\\
    \hline
    \end{tabular}
    }
    \caption{Effect on different backbone networks. The results of AP/APH on the Waymo Open validation set are reported. `*' represents that our DCLA and RWIoU-based regression loss are applied. }
\label{table2}
\end{table*}

%\begin{table}[t]
%    \centering
%    \begin{tabular}{c|ccc|ccc|ccc}
%    \hline
%    \multirow{2}{*}{Method} & \multicolumn{3}{c|}{Car} & \multicolumn{3}{c|}{Pedestrian} & \multicolumn{3}{c}{Cyclist} \\
%                 & Easy & Mod. & Hard & Easy & Mod. & Hard & Easy & Mod. & Hard\\
%    \hline\hline
%    PointPillars &88.59 &78.44 &75.73 &47.25 &41.90 &38.67 &82.10 &61.79 &57.83\\
%    PointPillars*&90.67 &79.86 &77.06 &48.65 &42.04 &38.17 &84.59 &64.46 &61.23\\
%    SECOND       &89.98 &81.56 &78.56 &52.02 &47.07 &42.07 &83.04 &67.67 &63.04\\
%    SECOND*      &91.73 &83.06 &82.28 &62.10 &55.12 &49.99 &91.63 &72.24 &67.70\\
%    PV-RCNN      &88.59 &78.44 &75.73 &47.25 &41.90 & 38.67& 82.10& 61.79& 57.83\\
%    PV-RCNN*     &88.59 &78.44 &75.73 &47.25 &41.90 & 38.67& 82.10& 61.79& 57.83\\
%   \hline
%    \end{tabular}
%    \label{table3}
%    \caption{Performance comparisons on the Waymo \textit{validation} set. The best results of detectors are highlighted in bold. "*" represents that the method is re-implemented using the official codebase OpenPCDet \cite{openpcdet}.}
%\end{table}

\subsection{Implementation Setup}

%\subsubsection{Dataset.}
%The Waymo Open dataset \cite{waymo} has a total of 798 training sequences with 158,361 LiDAR samples and 202 validation sequences with 40,077 LiDAR samples. The KITTI dataset \cite{kitti} contains 7,481 training samples and 7,518 test samples, where the training set is typically divided into the \textit{train} split with 3,712 samples and the \textit{val} split with 3,769 samples.

\subsubsection{Data Preprocessing} 
For the Waymo Open dataset, the detection range is $[-74.88, 74.88]m$ for the $X$ and $Y$ axes and $[-2, 4]m$ for the $Z$ axis, the voxel size is set to $(0.08, 0.08, 0.15)m$. For the KITTI dataset, the detection range is $[0, 70.4]m$ for the $X$ axis, $[-40, 40]m$ for the $Y$ axis, and $[-5, 3]m$ for the $Z$ axis, the voxel size is set to $(0.05, 0.05, 0.1)m$. 

%For the Waymo Open and KITTI datasets, the data augmentation schemes are kept the same as those of DSVT \cite{pv-rcnn} and PV-RCNN \cite{pv-rcnn}, respectively. 

\subsubsection{Training Details}
The backbone of our DCDet is the same as that of CenterPoint \cite{centerpoint}. Following PillarNeXt \cite{pillarnext}, we use a feature upsampling in the detection head of DCDet, which increase the output resolution with only a little overhead. All models are trained from scratch in an end-to-end manner with the Adam optimizer and a 0.003 learning rate. And the parameter $\alpha$ used in Eq.~(\ref{Eq:eq4}) is set to 0.5. The parameters $\lambda_{cls}$ and $\lambda_{iou}$ used in Eq.~(\ref{Eq:eq7}) are all set to 1. And the parameter $\lambda_{reg}$ used in Eq.~(\ref{Eq:eq1}) and Eq.~(\ref{Eq:eq7}) is set to 3. For the Waymo Open and KITTI datasets, the parameter $r$ used in DCLA is set to 1 and 3, respectively. On the Waymo Open and KITTI datasets, models are trained for 30 epochs with a batch size of 24 and 80 epochs with a batch size of 8, respectively. Hyper-parameters analysis is in Appendix. 

\subsection{Comparison with State-of-the-Art Methods}
The baseline models presented in Table~\ref{table1} primarily utilize either center-based or anchor-based label assignment. Moreover, they commonly employ $L_{norm}$ regression loss. As depicted in Table~\ref{table1}, the center-based label assignment demonstrates a significant advantage over the anchor-based label assignment on the Waymo Open dataset. Nevertheless, our DCDet, featuring a lightweight single-stage network, surpasses the state-of-the-art center-based method DSVT, which employs a heavy backbone network. Notably, even our DCDet model trained on only 20\% of the training samples outperforms both the center-based and anchor-based methods trained on the entire dataset. These results demonstrate the superior performance of our DCDet framework which employs DCLA and RWIoU-based regression loss.

%We also evaluate our DCDet on the test set of Waymo Open by submitting the results to the official server. The performance comparisons are shown in Table~\ref{table6}, our DCDet outperforms previous state-of-the-art methods significantly. Especially on small-scale categories i.e. pedestrian and cyclist, our method takes a significant lead thanks to the balanced and adequate positive samples provided by DCLA. 

We also evaluated our DCDet on the Waymo Open test set by submitting the results to the official server. The performance comparisons are presented in Table~\ref{table6}, revealing that our DCDet surpasses previous state-of-the-art methods significantly. Particularly, in the case of small-scale categories such as pedestrians and cyclists, our method demonstrates a substantial advantage due to the balanced and sufficient positive samples provided by DCLA.

\subsection{Effect on Different Backbone Networks} 

To assess the generality of our DCLA and RWIoU, we conduct experiments by incorporating them into several widely used backbone networks, namely SECOND, PillarNet, and DSVT. All models are reproduced using the OpenPCDet \cite{openpcdet} codebase. We train these models using both 20\% and 100\% of the training data from the Waymo Open dataset and present the results in Table~\ref{table2}. As evident from the table, the integration of our DCLA and RWIoU yields significant improvements across all model groups. This underscores the generality and effectiveness of our proposed DCLA and RWIoU techniques. Notably, the DCLA and RWIoU-based regression loss belong to the learning strategies of models, resulting in cost-free improvements. Even when trained on only 20\% of the training data, the models integrated with our DCLA and RWIoU techniques either surpass or catch up to the performance of models trained on the entire training data without these enhancements. This demonstrates that our learning strategies enhance the utilization of training data, which is particularly valuable considering the high cost associated with labeling 3D bounding boxes.

\subsection{Ablation Study}

%To further study the influence of each component of DCDet, we perform a comprehensive ablation analysis on the Waymo Open and KITTI datasets. For the Waymo Open dataset, models are trained on random 20\% training samples and evaluated on the whole validation samples. And for the KITTI dataset, models are trained on the \textit{train} set and evaluated on the \textit{val} set. 

To further study the influence of each component of DCDet, we perform a comprehensive ablation analysis on the Waymo Open and KITTI datasets. For the Waymo Open dataset, following prior works \cite{pv-rcnn,dsvt}, models are trained on 20\% training samples and evaluated on the whole validation samples. For the KITTI dataset, models are trained on the \textit{train} set and evaluated on the \textit{val} set.

\subsubsection{Effect of \textit{RWIoU} and \textit{DCLA}} 

The baseline model adopts center-based label assignment and $L_1$ regression loss. To evaluate the effectiveness of our proposed methods, we systematically integrate RWIoU-based regression loss and DCLA into the baseline model. The ablation results are presented in Table~\ref{table3}. We observe a notable performance improvement when incorporating RWIoU-based regression loss, as demonstrated by the results in the 1$^{st}$ and 2$^{nd}$ rows of Table~\ref{table3}. This suggests that the proposed loss function is better suited for the task of 3D object detection compared to the traditional $L_1$ loss. Furthermore, models trained with DCLA consistently achieve significantly better performance than the baseline, as illustrated in the 1$^{st}$ and 3$^{rd}$ rows of Table~\ref{table3}. This indicates that DCLA facilitates improved utilization of the available training data, thus enhancing the overall model performance. Notably, when both RWIoU-based regression loss and DCLA are used, the model achieves the highest performance among all evaluated models. These findings validate the effectiveness of our proposed methods and highlight the importance of carefully designing the loss function and label assignment for improving the performance of 3D object detectors.

\begin{table}[t]
    \centering
    %\resizebox{\linewidth}{!}{
    \begin{tabular}{ccccc}
        \hline
        \textit{RWIOU} & \textit{DCLA} & {Vehicle} & {Pedestrian} & {Cyclist}\\
        \hline
                     &            & 69.2/68.7 & 73.4/68.5 & 72.6/71.5 \\ 
        $\checkmark$ &            & 69.9/69.3 & 74.3/68.5 & 74.1/73.1 \\
                    &$\checkmark$ & 70.5/70.0 & 75.2/69.7 & 74.4/73.3 \\
        $\checkmark$&$\checkmark$&\textbf{71.0/70.5} & \textbf{75.9/70.1} & \textbf{75.1/74.0} \\
        \hline
    \end{tabular}%}
    \caption{Effect of different components of DCDet. \textit{RWIoU} and \textit{DCLA} denote RWIoU-based regression loss and dynamic cross label assignment, respectively. The LEVEL 2 AP/APH results on the Waymo Open validation set are reported.}
\label{table3}
\end{table}

\subsubsection{Comparison with Other Regression Losses} 
Table~\ref{table4} provides a comparison of different regression losses. All models utilize the DCLA scheme and the same backbone network. The results in the $1^{st}$, $2^{nd}$, and $3^{rd}$ rows of Table~\ref{table4} reveal marginal differences between the $L_1$, RDIoU-based \cite{rdiou}, and ODIoU-based \cite{pillarnet} regression losses. However, our RWIoU-based loss exhibits a significant performance improvement compared to the other regression losses, as demonstrated in the $4^{th}$ row of Table~\ref{table4}. These results highlight the effectiveness of our RWIoU, which decouples the rotation from IoU calculation by introducing rotation weighting. Notably, the RDIoU-based loss necessitates an additional direction classification loss, and the ODIoU-based loss requires an extra $L_1$ loss. In contrast, our RWIoU-based loss is a pure IoU-based loss without any auxiliary losses. This simplification allows our approach to achieve superior performance without introducing additional complexity.

\subsubsection{Comparison with Other Label Assignment Schemes} 
Table~\ref{table5} compares different label assignment schemes with all models using the RWIoU-based regression loss and the same backbone network. As depicted in the $1^{st}$ and $3^{rd}$ rows of Table~\ref{table5}, both anchor-based and box-based label assignment exhibit subpar performance when it comes to small objects like pedestrians and cyclists. This is mainly due to the unbalanced assignment of positive samples for objects with different scales. On the other hand, the center-based label assignment, as shown in the $2^{nd}$ row of Table~\ref{table5}, achieves good results on the Waymo Open dataset but performs poorly on the KITTI dataset. We argue that this discrepancy arises from overlooking a large number of excellent samples, resulting in an insufficient number of positive samples for training on small-scale datasets like KITTI. Moreover, the poor performance of simOTA \cite{ge2021yolox} in 3D object detection, as demonstrated in the $4^{th}$ row of Table~\ref{table5}, highlights the challenges of directly transferring methods from the 2D domain to the 3D domain. However, our DCLA outperforms these baseline label assignment schemes on both the Waymo Open and KITTI datasets, as illustrated in the last row of Table~\ref{table5}. This confirms that our DCLA can adapt to datasets of different scales by enabling balanced and adequate positive sampling.

\begin{table}[t]
    \centering
    %\resizebox{\linewidth}{!}{
    \begin{tabular}{cccc}
        \hline
        Regression Loss & {Vehicle} & {Pedestrian} & {Cyclist}\\
        \hline
        $L_1$          & 70.3/69.8 & 75.0/69.6 & 74.0/73.0 \\ 
        RDIoU-based & 70.2/69.7 & 74.8/69.3 & 74.3/73.2 \\
        ODIoU-based & 70.5/70.0 & 75.2/69.7 & 74.4/73.3 \\ 
        RWIoU-based & \textbf{71.0/70.5} & \textbf{75.9/70.1} & \textbf{75.1/74.0} \\
        %RWIoU-based* & \textbf{71.0/70.6} & 76.0/70.1 & 74.2/73.1 \\
        \hline
    \end{tabular}%}
    \caption{Comparison results of different regression losses. The LEVEL 2 AP/APH results on the Waymo Open validation set are reported.}
\label{table4}
\end{table}

\begin{table}[t]
    \centering
    \resizebox{\linewidth}{!}{
    \begin{tabular}{ccccc}
        \hline
        \multirow{2}{*}{Lable Assignment} & \multicolumn{3}{c}{Waymo} & KITTI\\
        \cline{2-4}
        & {Vehicle} & {Pedestrian} & {Cyclist} & Mod. Car\\
        \hline
        Anchor-based     & 67.8/67.3 & 63.4/55.5 & 67.7/66.5 & 85.37\\ 
        Center-based     & 69.9/69.3 & 74.3/68.5 & 74.1/73.1 & 75.49\\
        Box-based        & 67.8/67.4 & 66.2/61.4 & 69.9/69.0 & 85.32\\
        simOTA           & 68.7/68.3 & 67.8/63.1 & 72.2/71.2 & 85.45\\
        DCLA             & \textbf{71.0/70.5} & \textbf{75.9/70.1} & \textbf{75.1/74.0} & \textbf{85.82}\\
        \hline
    \end{tabular}}
    \caption{Comparison results of different label assignment schemes. The LEVEL 2 AP/APH results on the Waymo Open validation set and moderate AP$_{R40}$ results on the KITTI \textit{val} are reported. }
\label{table5}
\end{table}

\section{Conclusion}
In this paper, we propose a dynamic cross label assignment (DCLA), which dynamically assigns positive samples from a cross-shaped region for each object. The DCLA scheme mitigates the imbalanced issue in the anchor-based assignment and the loss of high-quality samples in the center-based assignment. Thanks to the balanced and adequate positive sampling, DCLA effectively adapts to different scale datasets. Moreover, a rotation-weighted IoU (RWIoU), which considers the rotation and direction in a weighting way, is introduced to measure the proximity of two rotation boxes. Extensive experiments conducted on various datasets demonstrate the generality and effectiveness of our methods. 

\section*{Acknowledgments}
\noindent This work is supported by the Project of Guangxi Key R \& D Program (No. GuikeAB24010324).

\bibliographystyle{named}
\bibliography{ijcai24}

\begin{thebibliography}{}

\bibitem[\protect\citeauthoryear{Deng \bgroup \em et al.\egroup
  }{2021}]{voxel-rcnn}
Jiajun Deng, Shaoshuai Shi, Peiwei Li, Wengang Zhou, Yanyong Zhang, and
  Houqiang Li.
\newblock Voxel r-cnn: Towards high performance voxel-based 3d object
  detection.
\newblock In {\em AAAI}, 2021.

\bibitem[\protect\citeauthoryear{Fan \bgroup \em et al.\egroup }{2022a}]{sst}
Lue Fan, Ziqi Pang, Tianyuan Zhang, Yu-Xiong Wang, Hang Zhao, Feng Wang, Naiyan
  Wang, and Zhaoxiang Zhang.
\newblock Embracing single stride 3d object detector with sparse transformer.
\newblock In {\em CVPR}, 2022.

\bibitem[\protect\citeauthoryear{Fan \bgroup \em et al.\egroup }{2022b}]{fsd}
Lue Fan, Feng Wang, Naiyan Wang, and ZHAO-XIANG ZHANG.
\newblock Fully sparse 3d object detection.
\newblock In {\em NeurIPS}, 2022.

\bibitem[\protect\citeauthoryear{Ge \bgroup \em et al.\egroup }{2020}]{afdet}
Runzhou Ge, Zhuangzhuang Ding, Yihan Hu, Yu~Wang, Sijia Chen, Li~Huang, and
  Yuan Li.
\newblock Afdet: Anchor free one stage 3d object detection.
\newblock {\em arXiv preprint arXiv:2006.12671}, 2020.

\bibitem[\protect\citeauthoryear{Ge \bgroup \em et al.\egroup
  }{2021}]{ge2021yolox}
Zheng Ge, Songtao Liu, Feng Wang, Zeming Li, and Jian Sun.
\newblock Yolox: Exceeding yolo series in 2021.
\newblock {\em arXiv preprint arXiv:2107.08430}, 2021.

\bibitem[\protect\citeauthoryear{Geiger \bgroup \em et al.\egroup
  }{2012}]{kitti}
Andreas Geiger, Philip Lenz, and Raquel Urtasun.
\newblock Are we ready for autonomous driving? the kitti vision benchmark
  suite.
\newblock In {\em CVPR}, 2012.

\bibitem[\protect\citeauthoryear{He \bgroup \em et al.\egroup }{2022}]{voxset}
Chenhang He, Ruihuang Li, Shuai Li, and Lei Zhang.
\newblock Voxel set transformer: A set-to-set approach to 3d object detection
  from point clouds.
\newblock In {\em CVPR}, 2022.

\bibitem[\protect\citeauthoryear{Hu \bgroup \em et al.\egroup }{2022}]{afdetv2}
Yihan Hu, Zhuangzhuang Ding, Runzhou Ge, Wenxin Shao, Li~Huang, Kun Li, and
  Qiang Liu.
\newblock Afdetv2: Rethinking the necessity of the second stage for object
  detection from point clouds.
\newblock In {\em AAAI}, 2022.

\bibitem[\protect\citeauthoryear{Lang \bgroup \em et al.\egroup
  }{2019}]{pointpillars}
Alex~H Lang, Sourabh Vora, Holger Caesar, Lubing Zhou, Jiong Yang, and Oscar
  Beijbom.
\newblock Pointpillars: Fast encoders for object detection from point clouds.
\newblock In {\em CVPR}, 2019.

\bibitem[\protect\citeauthoryear{Li \bgroup \em et al.\egroup
  }{2021}]{lidarrcnn}
Zhichao Li, Feng Wang, and Naiyan Wang.
\newblock Lidar r-cnn: An efficient and universal 3d object detector.
\newblock In {\em CVPR}, 2021.

\bibitem[\protect\citeauthoryear{Li \bgroup \em et al.\egroup
  }{2023}]{pillarnext}
Jinyu Li, Chenxu Luo, and Xiaodong Yang.
\newblock Pillarnext: Rethinking network designs for 3d object detection in
  lidar point clouds.
\newblock In {\em CVPR}, 2023.

\bibitem[\protect\citeauthoryear{Lin \bgroup \em et al.\egroup
  }{2017}]{focalloss}
Tsung-Yi Lin, Priya Goyal, Ross Girshick, Kaiming He, and Piotr Doll{\'a}r.
\newblock Focal loss for dense object detection.
\newblock In {\em ICCV}, 2017.

\bibitem[\protect\citeauthoryear{Qi \bgroup \em et al.\egroup
  }{2017a}]{pointnet}
Charles~R Qi, Hao Su, Kaichun Mo, and Leonidas~J Guibas.
\newblock Pointnet: Deep learning on point sets for 3d classification and
  segmentation.
\newblock In {\em CVPR}, 2017.

\bibitem[\protect\citeauthoryear{Qi \bgroup \em et al.\egroup
  }{2017b}]{pointnet++}
Charles~Ruizhongtai Qi, Li~Yi, Hao Su, and Leonidas~J Guibas.
\newblock Pointnet++: Deep hierarchical feature learning on point sets in a
  metric space.
\newblock In {\em NeurIPS}, 2017.

\bibitem[\protect\citeauthoryear{Rezatofighi \bgroup \em et al.\egroup
  }{2019}]{GIoU}
Hamid Rezatofighi, Nathan Tsoi, JunYoung Gwak, Amir Sadeghian, Ian Reid, and
  Silvio Savarese.
\newblock Generalized intersection over union: A metric and a loss for bounding
  box regression.
\newblock In {\em CVPR}, 2019.

\bibitem[\protect\citeauthoryear{Sheng \bgroup \em et al.\egroup
  }{2022}]{rdiou}
Hualian Sheng, Sijia Cai, Na~Zhao, Bing Deng, Jianqiang Huang, Xian-Sheng Hua,
  Min-Jian Zhao, and Gim~Hee Lee.
\newblock Rethinking iou-based optimization for single-stage 3d object
  detection.
\newblock In {\em ECCV}, 2022.

\bibitem[\protect\citeauthoryear{Shi \bgroup \em et al.\egroup
  }{2019}]{pointrcnn}
Shaoshuai Shi, Xiaogang Wang, and Hongsheng Li.
\newblock Pointrcnn: 3d object proposal generation and detection from point
  cloud.
\newblock In {\em CVPR}, 2019.

\bibitem[\protect\citeauthoryear{Shi \bgroup \em et al.\egroup
  }{2020a}]{pv-rcnn}
Shaoshuai Shi, Chaoxu Guo, Li~Jiang, Zhe Wang, Jianping Shi, Xiaogang Wang, and
  Hongsheng Li.
\newblock Pv-rcnn: Point-voxel feature set abstraction for 3d object detection.
\newblock In {\em CVPR}, 2020.

\bibitem[\protect\citeauthoryear{Shi \bgroup \em et al.\egroup
  }{2020b}]{parta2}
Shaoshuai Shi, Zhe Wang, Jianping Shi, Xiaogang Wang, and Hongsheng Li.
\newblock From points to parts: 3d object detection from point cloud with
  part-aware and part-aggregation network.
\newblock {\em TPAMI}, 2020.

\bibitem[\protect\citeauthoryear{Shi \bgroup \em et al.\egroup
  }{2022}]{pillarnet}
Guangsheng Shi, Ruifeng Li, and Chao Ma.
\newblock Pillarnet: Real-time and high-performance pillar-based 3d object
  detection.
\newblock In {\em ECCV}, 2022.

\bibitem[\protect\citeauthoryear{Shi \bgroup \em et al.\egroup
  }{2023}]{pv-rcnn++}
Shaoshuai Shi, Li~Jiang, Jiajun Deng, Zhe Wang, Chaoxu Guo, Jianping Shi,
  Xiaogang Wang, and Hongsheng Li.
\newblock Pv-rcnn++: Point-voxel feature set abstraction with local vector
  representation for 3d object detection.
\newblock {\em IJCV}, 2023.

\bibitem[\protect\citeauthoryear{Sun \bgroup \em et al.\egroup }{2020}]{waymo}
Pei Sun, Henrik Kretzschmar, Xerxes Dotiwalla, Aurelien Chouard, Vijaysai
  Patnaik, Paul Tsui, James Guo, Yin Zhou, Yuning Chai, Benjamin Caine, et~al.
\newblock Scalability in perception for autonomous driving: Waymo open dataset.
\newblock In {\em CVPR}, 2020.

\bibitem[\protect\citeauthoryear{Sun \bgroup \em et al.\egroup
  }{2022}]{swformer}
Pei Sun, Mingxing Tan, Weiyue Wang, Chenxi Liu, Fei Xia, Zhaoqi Leng, and
  Dragomir Anguelov.
\newblock Swformer: Sparse window transformer for 3d object detection in point
  clouds.
\newblock In {\em ECCV}, 2022.

\bibitem[\protect\citeauthoryear{Team}{2020}]{openpcdet}
OpenPCDet~Development Team.
\newblock Openpcdet: An open-source toolbox for 3d object detection from point
  clouds.
\newblock \url{https://github.com/open-mmlab/OpenPCDet}, 2020.

\bibitem[\protect\citeauthoryear{Tian \bgroup \em et al.\egroup }{2019}]{fcos}
Zhi Tian, Chunhua Shen, Hao Chen, and Tong He.
\newblock Fcos: Fully convolutional one-stage object detection.
\newblock In {\em ICCV}, 2019.

\bibitem[\protect\citeauthoryear{Wang \bgroup \em et al.\egroup
  }{2021}]{3dcenternet}
Qi~Wang, Jian Chen, Jianqiang Deng, and Xinfang Zhang.
\newblock 3d-centernet: 3d object detection network for point clouds with
  center estimation priority.
\newblock {\em Pattern Recognition}, 2021.

\bibitem[\protect\citeauthoryear{Wang \bgroup \em et al.\egroup }{2023}]{dsvt}
Haiyang Wang, Chen Shi, Shaoshuai Shi, Meng Lei, Sen Wang, Di~He, Bernt
  Schiele, and Liwei Wang.
\newblock Dsvt: Dynamic sparse voxel transformer with rotated sets.
\newblock In {\em CVPR}, 2023.

\bibitem[\protect\citeauthoryear{Xu \bgroup \em et al.\egroup }{2022}]{btcdet}
Qiangeng Xu, Yiqi Zhong, and Ulrich Neumann.
\newblock Behind the curtain: Learning occluded shapes for 3d object detection.
\newblock In {\em AAAI}, 2022.

\bibitem[\protect\citeauthoryear{Yan \bgroup \em et al.\egroup }{2018}]{second}
Yan Yan, Yuxing Mao, and Bo~Li.
\newblock Second: Sparsely embedded convolutional detection.
\newblock {\em Sensors}, 2018.

\bibitem[\protect\citeauthoryear{Yang \bgroup \em et al.\egroup }{2020}]{3dssd}
Zetong Yang, Yanan Sun, Shu Liu, and Jiaya Jia.
\newblock 3dssd: Point-based 3d single stage object detector.
\newblock In {\em CVPR}, 2020.

\bibitem[\protect\citeauthoryear{Yin \bgroup \em et al.\egroup
  }{2021}]{centerpoint}
Tianwei Yin, Xingyi Zhou, and Philipp Krahenbuhl.
\newblock Center-based 3d object detection and tracking.
\newblock In {\em CVPR}, 2021.

\bibitem[\protect\citeauthoryear{Zhang \bgroup \em et al.\egroup }{2020}]{atss}
Shifeng Zhang, Cheng Chi, Yongqiang Yao, Zhen Lei, and Stan~Z Li.
\newblock Bridging the gap between anchor-based and anchor-free detection via
  adaptive training sample selection.
\newblock In {\em CVPR}, 2020.

\bibitem[\protect\citeauthoryear{Zhang \bgroup \em et al.\egroup
  }{2022a}]{eiou}
Yi-Fan Zhang, Weiqiang Ren, Zhang Zhang, Zhen Jia, Liang Wang, and Tieniu Tan.
\newblock Focal and efficient iou loss for accurate bounding box regression.
\newblock {\em Neurocomputing}, 2022.

\bibitem[\protect\citeauthoryear{Zhang \bgroup \em et al.\egroup
  }{2022b}]{iassd}
Yifan Zhang, Qingyong Hu, Guoquan Xu, Yanxin Ma, Jianwei Wan, and Yulan Guo.
\newblock Not all points are equal: Learning highly efficient point-based
  detectors for 3d lidar point clouds.
\newblock In {\em CVPR}, 2022.

\bibitem[\protect\citeauthoryear{Zheng \bgroup \em et al.\egroup }{2020}]{DIoU}
Zhaohui Zheng, Ping Wang, Wei Liu, Jinze Li, Rongguang Ye, and Dongwei Ren.
\newblock Distance-iou loss: Faster and better learning for bounding box
  regression.
\newblock In {\em AAAI}, 2020.

\bibitem[\protect\citeauthoryear{Zheng \bgroup \em et al.\egroup
  }{2021}]{cia-ssd}
Wu~Zheng, Weiliang Tang, Sijin Chen, Li~Jiang, and Chi-Wing Fu.
\newblock Cia-ssd: Confident iou-aware single-stage object detector from point
  cloud.
\newblock In {\em AAAI}, 2021.

\bibitem[\protect\citeauthoryear{Zhou and Tuzel}{2018}]{voxelnet}
Yin Zhou and Oncel Tuzel.
\newblock Voxelnet: End-to-end learning for point cloud based 3d object
  detection.
\newblock In {\em CVPR}, 2018.

\bibitem[\protect\citeauthoryear{Zhou \bgroup \em et al.\egroup
  }{2019a}]{3diou}
Dingfu Zhou, Jin Fang, Xibin Song, Chenye Guan, Junbo Yin, Yuchao Dai, and
  Ruigang Yang.
\newblock Iou loss for 2d/3d object detection.
\newblock In {\em 3DV}, 2019.

\bibitem[\protect\citeauthoryear{Zhou \bgroup \em et al.\egroup
  }{2019b}]{centernet}
Xingyi Zhou, Dequan Wang, and Philipp Kr{\"a}henb{\"u}hl.
\newblock Objects as points.
\newblock {\em arXiv preprint arXiv:1904.07850}, 2019.

\bibitem[\protect\citeauthoryear{Zhou \bgroup \em et al.\egroup
  }{2022}]{centerformer}
Zixiang Zhou, Xiangchen Zhao, Yu~Wang, Panqu Wang, and Hassan Foroosh.
\newblock Centerformer: Center-based transformer for 3d object detection.
\newblock In {\em ECCV}, 2022.

\bibitem[\protect\citeauthoryear{Zhu \bgroup \em et al.\egroup
  }{2020}]{zhu2020autoassign}
Benjin Zhu, Jianfeng Wang, Zhengkai Jiang, Fuhang Zong, Songtao Liu, Zeming Li,
  and Jian Sun.
\newblock Autoassign: Differentiable label assignment for dense object
  detection.
\newblock {\em arXiv preprint arXiv:2007.03496}, 2020.

\end{thebibliography}

\clearpage

\appendix

\begin{table}[t]
    \centering
    %\resizebox{\linewidth}{!}{
    \begin{tabular}{cccc}
        \hline
        $\alpha$ & {Vehicle} & {Pedestrian} & {Cyclist}\\
        \hline 
        1.00          & 71.0/70.5 & 75.4/69.9 & 74.6/73.5 \\ 
        0.75          & 70.9/70.4 & 75.6/70.0 & 74.6/73.5 \\
        0.50          & 71.0/70.5 & 75.9/70.1 & 75.1/74.0 \\
        0.25          & 70.9/70.4 & 75.8/70.1 & 74.7/73.6 \\
        \hline
    \end{tabular}%}
    \caption{Effect of different $\alpha$ settings. Models are trained on 20\% training samples of the Waymo Open dataset. The LEVEL 2 AP/APH results are reported.}
\label{tableA1}
\end{table}

\section{Gradient Analysis of RWIoU}
For a given predicted box $\textbf{b}_p = \{x_p, y_p, z_p, l_p, w_p, h_p, s_p, c_p\}$ and its ground truth box $\textbf{b}_t = \{x_t, y_t, z_t, l_t, w_t, h_t, s_t, c_t\}$. $(x, y, z)$ denotes the center location of a 3D bounding box. $(l, w, h)$ are the length, width, and height of a 3D bounding box, respectively. $(s, c)$ are the sine and cosine values of the orientation of a 3D bounding box. The RWIoU loss is calculated as follows: 

\begin{equation}
    \begin{aligned}
    L_{rwiou} &= 1 - RWIoU(\textbf{b}_t, \textbf{b}_p), \\
              &= 1 - \frac{V_{weighted}}{V_{union}},
    \end{aligned}
    \label{Eq:eqA1}
\end{equation}

\noindent where $V_{weighted}$ and $V_{union}$ are calculated as in Eq.~(\ref{Eq:eq5}) and Eq.~(\ref{Eq:eq6}), respectively. To analyze the gradient of the RWIoU loss, we need to calculate the partial derivatives of RWIoU loss w.r.t. the attributes of the 3D bounding box.   

First, we calculate the partial derivative of RWIoU loss w.r.t. $s_p$ as follows:

\begin{equation}
    \begin{aligned}
    \frac{\partial L_{rwiou}}{\partial s_p} &= \frac{\frac{\partial V_{union}}{\partial s_p}V_{weighted} - \frac{\partial V_{weighted}}{\partial s_p}V_{union}}{ V_{union}^2},\\
    &= - \frac{\frac{\partial V_{weighted}}{\partial s_p} (V_{weighted} + V_{union})}{V_{union}^2},\\
    &= - \frac{\frac{\partial \omega_{s}}{\partial s_p} \omega_{c} V_{inter}(V_{weighted} + V_{union})}{V_{union}^2},\\
    &= -\frac{\partial \omega_s}{\partial s_p} \omega_{c} (RWIoU + 1) \frac{V_{inter}}{V_{union}}, \\
    &= 
        \begin{cases}
            \frac{\alpha}{2} \omega_{c} (RWIoU + 1) \frac{V_{inter}}{V_{union}},  \text{ if } s_p > s_t, \\
            - \frac{\alpha}{2} \omega_{c} (RWIoU + 1) \frac{V_{inter}}{V_{union}}, \text{ if } s_p < s_t,
        \end{cases}
    \end{aligned}
    \label{Eq:eqA2}
\end{equation}

\noindent where $V_{inter}$ is calculated as in Eq.~(\ref{Eq:eq4}), $\omega_c \in \left[0,1\right]$, $RWIoU \in \left[0,1\right]$, and $\frac{V_{inter}}{V_{union}} \in \left[0,1\right]$. Therefore, the gradient $|\frac{\partial L_{rwiou}}{\partial s_p}| \in [0, \alpha]$. The same reasoning leads to the partial derivative of RWIoU loss w.r.t. $c_p$. 

Then, we calculate the partial derivative of RWIoU loss w.r.t. center location. There are too many cases for the calculation of $V_{inter}$. Here, we only consider the case as shown in Figure~\ref{fig3} where the orange box is considered as the predicted box. Thus, we get the partial derivative of RWIoU loss w.r.t. $x_p$ as follows:

\begin{equation}
    \begin{aligned}
    \frac{\partial L_{rwiou}}{\partial x_p} &= \frac{\frac{\partial V_{union}}{\partial x_p}V_{weighted} - \frac{\partial V_{weighted}}{\partial x_p}V_{union}}{ V_{union}^2},\\
    &= - \frac{\partial V_{inter}}{\partial x_p}  \cdot \frac{\omega(V_{weighted} + V_{union})}{V_{union}^2},\\
    \end{aligned}
    \label{Eq:eqA3}
\end{equation}

\begin{equation}
    \begin{aligned}
    \frac{\partial V_{inter}}{\partial x_p} = (S_T - S_B)(S_U - S_D),\\
    \end{aligned}
\label{Eq:eqA4}
\end{equation}

\noindent where $S_T$, $S_B$, $S_U$ and $S_D$ are calculated in Eq.~(\ref{Eq:eq4}), and $\omega$ is calculated in Eq.~(\ref{Eq:eq5}). The same reasoning leads to the partial derivatives of RWIoU loss w.r.t. $y_p$ and $z_p$. According to the Eq.~(\ref{Eq:eqA3}) and Eq.~(\ref{Eq:eqA4}), we can conclude that the gradient $|\frac{\partial L_{rwiou}}{\partial x_p}|$ will be increased as the converge of the model. But there is an upper bound $\frac{2}{l_t}$, when $\textbf{b}_p$ is infinitely close to $\textbf{b}_t$. 

Next, we calculate the partial derivative of RWIoU loss w.r.t. scale. Generally, the center locations of $\textbf{b}_p$ and $\textbf{b}_t$ are very close. For simplicity, we consider the case that the center locations of the two boxes are well aligned. Thus, we obtain the partial derivative of RWIoU loss w.r.t. $l_p$ as follows:

\begin{equation}
    \begin{aligned}
    \frac{\partial L_{rwiou}}{\partial l_p} =& \frac{\frac{\partial V_{union}}{\partial l_p}V_{weighted} - \frac{\partial V_{weighted}}{\partial l_p}V_{union}}{ V_{union}^2},\\
     = &
    \begin{cases}
        - RWIoU \frac{\omega V_t}{l_p V_{union}},  \text{ if } l_p < l_t, \\
         RWIoU  \frac{V_p}{l_p V_{union}}, \text{ if } l_p > l_t,
    \end{cases}
    \end{aligned}
\label{Eq:eqA5}
\end{equation}

\noindent where $V_p$ and $V_t$ are the volume of $\textbf{b}_p$ and $\textbf{b}_t$, respectively. The same reasoning leads to the partial derivatives of RWIoU loss w.r.t. $w_p$ and $h_p$. According to the Eq.~(\ref{Eq:eqA5}), we can conclude that the gradient $|\frac{\partial L_{rwiou}}{\partial l_p}|$ will be increased as the converge of the model. But there is an upper bound $\frac{1}{l_t}$, when $\textbf{b}_p$ is infinitely close to $\textbf{b}_t$. 

\begin{table}[t]
    \centering
    %\resizebox{\linewidth}{!}{
    \begin{tabular}{cccc}
        \hline
        $\lambda_{reg}$ & {Vehicle} & {Pedestrian} & {Cyclist}\\
        \hline
        1          & 70.9/70.5 & 75.6/70.0 & 74.0/72.9 \\ 
        2          & 70.8/70.4 & 75.6/69.8 & 74.6/73.5 \\
        3          & 71.0/70.5 & 75.9/70.1 & 75.1/74.0 \\
        4          & 71.0/70.5 & 75.2/69.7 & 73.8/72.7 \\
        \hline
    \end{tabular}%}
    \caption{Effect of different $\lambda_{reg}$ setting. Models are trained on 20\% training samples of the Waymo Open dataset. The LEVEL 2 AP/APH results are reported.}
\label{tableA2}
\end{table}

\begin{table}[t]
    \centering
    %\resizebox{\linewidth}{!}{
    \begin{tabular}{cccc}
        \hline
        $r$ & {Vehicle} & {Pedestrian} & {Cyclist}\\
        \hline
        0          & 70.2/69.7 & 74.8/69.5 & 73.9/72.9 \\
        1          & 71.0/70.5 & 75.9/70.1 & 75.1/74.0 \\
        2          & 70.5/70.0 & 75.2/69.4 & 74.6/73.5 \\
        3          & 69.9/69.4 & 72.1/66.9 & 73.7/72.7 \\
        \hline
    \end{tabular}%}
    \caption{Effect of different $r$ setting. Models are trained on 20\% training samples of the Waymo Open dataset. The LEVEL 2 AP/APH results are reported.}
\label{tableA3}
\end{table}

\section{Hyper-parameters Analysis}
In this section, we determine the suitable values for the parameter $\alpha$ in Eq.\ref{Eq:eq5} and the regression loss weight $\lambda_{reg}$ through experiments conducted on the Waymo Open dataset. The performance of different $\alpha$ settings is presented in Table\ref{tableA1}, revealing minimal variations in performance across the different settings. However, when $\alpha = 0.5$, there is a slightly better performance compared to other settings. Similarly, Table~\ref{tableA2} showcases the performance comparisons of various $\lambda_{reg}$ settings, with minor differences observed between them. Notably, the best performance is achieved when $\lambda_{reg} = 3$. We also compare the performances with different $r$ settings. As shown in Table~\ref{tableA3}, the performance achieves the best when $r = 1$. Consequently, we adopt $\alpha = 0.5$, $\lambda_{reg} = 3$ and $r = 1$ as the default settings.
\end{document}